\documentclass[sigconf]{acmart}
\acmSubmissionID{1633}

\usepackage{balance}

\usepackage{algorithm}
\usepackage{algorithmic}
\usepackage{graphicx}
\usepackage{subfigure}
\usepackage{booktabs}
\usepackage{flushend}

\AtBeginDocument{%
  \providecommand\BibTeX{{%
    \normalfont B\kern-0.5em{\scshape i\kern-0.25em b}\kern-0.8em\TeX}}}

\setcopyright{acmcopyright}
\copyrightyear{2019} 
\acmYear{2019} 
\acmConference[MM '19]{Proceedings of the 27th ACM International Conference on Multimedia}{October 21--25, 2019}{Nice, France}
\acmBooktitle{Proceedings of the 27th ACM International Conference on Multimedia (MM '19), October 21--25, 2019, Nice, France}
\acmPrice{15.00}
\acmDOI{10.1145/3343031.3350571}
\acmISBN{978-1-4503-6889-6/19/10}

\begin{document}

%
\title{Neural Storyboard Artist: \\Visualizing Stories with Coherent Image Sequences}
%

\author{Shizhe Chen}
\affiliation{Renmin University of China}
\email{cszhe1@ruc.edu.cn}

\author{Bei Liu}
\affiliation{Microsoft Research Asia}
\email{bei.liu@microsoft.com}

\author{Jianlong Fu}
\affiliation{Microsoft Research Asia}
\email{jianf@microsoft.com}

\author{Ruihua Song}
\authornotemark[1]
\affiliation{Microsoft XiaoIce}
\email{rsong@microsoft.com}

\author{Qin Jin}
\authornote{Qin Jin and Ruihua Song are the corresponding authors.}
\affiliation{Renmin University of China}
\email{qjin@ruc.edu.cn}

\author{Pingping Lin}
\affiliation{Microsoft XiaoIce}
\email{pingping.lin@microsoft.com}

\author{Xiaoyu Qi}
\affiliation{Microsoft XiaoIce}
\email{xiaoyu.qi@microsoft.com}

\author{Chunting Wang}
\affiliation{Beijing Film Academy}
\email{04172154@mail.bfa.edu.cn}

\author{Jin Zhou}
\affiliation{Beijing Film Academy}
\email{whitezj@vip.sina.com}


%
\renewcommand{\shortauthors}{Chen, et al.}

%
\begin{abstract}
A storyboard is a sequence of images to illustrate a story containing multiple sentences, which has been a key process to create different story products.
In this paper, we tackle a new multimedia task of automatic storyboard creation to facilitate this process and inspire human artists.
Inspired by the fact that our understanding of languages is based on our past experience, we propose a novel inspire-and-create framework with a story-to-image retriever that selects relevant cinematic images for inspiration and a storyboard creator that further refines and renders images to improve the relevancy and visual consistency.
The proposed retriever dynamically employs contextual information in the story with hierarchical attentions and applies dense visual-semantic matching to accurately retrieve and ground images.
The creator then employs three rendering steps to increase the flexibility of retrieved images, which include erasing irrelevant regions, unifying styles of images and substituting consistent characters. 
We carry out extensive experiments on both in-domain and out-of-domain visual story datasets.
The proposed model achieves better quantitative performance than the state-of-the-art baselines for storyboard creation.
Qualitative visualizations and user studies further verify that our approach can create high-quality storyboards even for stories in the wild.
\end{abstract}

%
%


%

%
\maketitle

\section{Introduction}
A storyboard is a sequence of images to visualize a story with multiple sentences, which vividly conveys the story content shot by shot.
The storyboarding process is one of the most important stages to create different story products such as movie, animation etc.
It not only simplifies the understanding of textual stories with visual aids, but also makes following steps in story production go more smoothly via planning key images in advance. 
In this research, we explore how to automatically create a storyboard given a story.

Storyboard creation is nevertheless a challenging task even for professional artists, which requires many factors taken into consideration.
Firstly, images in professional storyboards are supposed to be \emph{\textbf{cinematic}} considering the framing, structure, view and so on.
Secondly, the visualized image should contain sufficient \emph{\textbf{relevant}} details to convey the story such as scenes, characters, actions etc.
Last but not least, the storyboard should look visually \emph{\textbf{consistent}} with coherent styles and characters across all images.

\begin{figure*}
\vspace{-5mm}
	\begin{center}
		\includegraphics[width=0.98\linewidth]{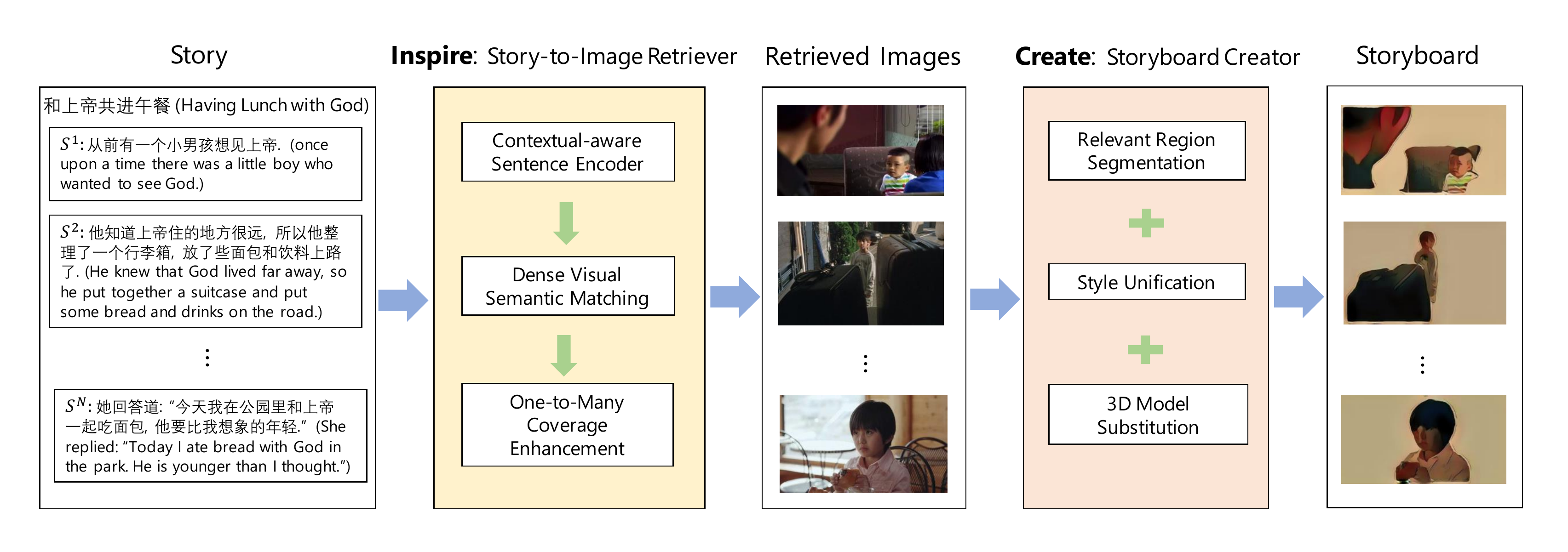}
	\end{center}
	\vspace{-7mm}
	\caption{The proposed inspire-and-create framework for storyboard creation: a story-to-image retriever is firstly utilized to retrieve relevant image sequences from cinematic image set for inspiration; then a storyboard creator simulates the retrieved images to create more flexible story visualizations.
	} 
	\label{fig:method_framework}
\vspace{-3mm}
\end{figure*}

There are mainly two directions in the literature that can be adopted to address the automatic storyboard creation task, namely generation-based and retrieval-based methods.
The generation-based method directly generates images conditioned on texts \cite{reed2016generative,zhang2017stackgan,xu2018attngan,pan2017create,li2018storygan} by generative adversarial learning (GAN), which is flexible to generate novel images.
However, it suffers from generating high-quality, diverse and relevant images due to the well-known training difficulties \cite{goodfellow2014generative,salimans2016improved,pan2017create,li2018storygan}.
The retrieval-based method overcomes the difficulty of image generation via retrieving existing high-quality images with texts \cite{kiros2014unifying,faghri2018vse++,nam2017dual,karpathy2015deep,ravi2018show,sun2019videobert}.
Nevertheless, current retrieval-based methods mainly contain three limitations for storyboard creation.
Firstly, most previous endeavors utilize single sentence to retrieve images without considering context. However, the story context plays an important role in understanding the constituent sentence and maintaining semantic coherence of retrieved images. 
Secondly, the retrieval-based method lacks flexibility since it cannot guarantee an existing image is precisely relevant to a novel input story. 
Thirdly, retrieved images can be extracted from different sources that makes the whole storyboard sequence visually inconsistent in styles and characters.

To overcome limitations of both generation- and retrieval-based methods, we propose a novel inspire-and-create framework for automatic storyboard creation.
Motivated by the embodied simulation hypothesis \cite{lakoff2012explaining} in 
psychology which considers human's understanding of language as the simulation of past experience about vision, sound etc., the proposed framework first takes inspirations from existing well-designed visual images via a story-to-image retriever, and then simulates the acquired images to create flexible story visualizations by a novel storyboard creator.
Figure~\ref{fig:method_framework} illustrates the overall structure of the inspire-and-create framework.
To be specific, we propose a contextual-aware dense visual-semantic matching model as the story-to-image retriever, which has three special characteristics for storyboard inspiration compared with previous retrieval works \cite{kiros2014unifying,faghri2018vse++,nam2017dual,karpathy2015deep,ravi2018show,sun2019videobert}: 
1) dynamically encode sentences with relevant story contexts; 2) ground each word on image regions to enable future rendering; 3) visualize one sentence with multiple images if necessary.
For the storyboard creator, we propose three steps for image rendering, including relevant region segmentation to erase irrelevant parts in the original retrieved image, style unification and 3D character substitution to improve the visual consistency on style and characters respectively.
Extensive experimental results on in-domain and out-of-domain datasets demonstrate the effectiveness of the proposed inspire-and-create model.
Our approach achieves better performance on both objective metrics and subjective human evaluations than the state-of-the-art retrieval based methods for storyboard creation.

The main contribution of this paper are as follows:
\begin{itemize}
	\item We propose a novel inspire-and-create framework for the challenging storyboard creation task. To the best of our knowledge, this is the first work focusing on automatic storyboard creation for stories in the wild.
	\item We propose a contextual-aware dense visual-semantic matching model as story-to-image retriever for inspiration, which not only achieves accurate retrieval but also enables one sentence visualized with multiple complementary images.
    \item The proposed storyboard creator consists of three rendering steps to simulate the retrieved images, which overcomes the inflexibility in retrieval based models and improves relevancy and visual consistency of generated storyboards.
	\item Both objective and subjective evaluations show that our approach obtains significant improvements over previous methods. Our proposed model can create cinematic, relevant and consistent storyboard even for out-of-domain stories.
\end{itemize}

\section{Related Work}
Previous works for texts visualization can be broadly divided into two types, which are generation-based and retrieval-based methods.

\subsection{Generation-based Methods}
Generation-based methods \cite{goodfellow2014generative} have the flexibility to generate novel outputs, which have been exploited in different tasks such as text generation \cite{liu2018beyond,li2019emotion}, image generation \cite{ma2018gan} and so on.
Recent works also explore to generate images conditioning on input texts \cite{reed2016generative,zhang2018photographic,pan2017create,li2018storygan}.
Most of them focus on single sentence to single image generation \cite{reed2016generative,zhang2017stackgan,xu2018attngan}.
Reed \emph{et al.} \cite{reed2016generative} propose to use conditional GAN with adversarial training of a generator and a discriminator to improve text-to-image generation ability.
Zhang \emph{et al.} \cite{zhang2017stackgan} propose Stacked GAN to generate larger size images via a sketch-refinement process in two stages.
Xu \emph{et al.} \cite{xu2018attngan} employ attention mechanism to attend on relevant words when synthesizing different regions of the image.
Recently, some endeavors have been put for generating image sequences given texts.
Pan \emph{et al.} \cite{pan2017create} utilize GAN to create a short video based on a single sentence, which improves motion smoothness of consecutive frames.
The storyboard creation, however, is different from short video generation, which emphasizes more on semantic coherency rather than low-level motion smoothness since a story can contain different scene changes.
To tackle such challenge, Li \emph{et al.} \cite{li2018storygan} propose StoryGAN for story visualization which employs a context encoder to track the story flow and two discriminators at the story and image level to enhance quality and consistency of generated images.
However, due to the well-known difficulties of training generative models \cite{goodfellow2014generative,salimans2016improved}, these works are limited on specific domains such as birds \cite{zhang2017stackgan}, flowers \cite{xu2018attngan}, numbers \cite{pan2017create} and cartoon characters \cite{li2018storygan} image generation where the structures are much easier, and the quality of generated image is usually unstable.
Therefore, it is hard to directly apply generative models in complex scenarios such storyboard creation for stories in the wild.

\subsection{Retrieval-based Methods}
Retrieval-based methods ensure to generate high-quality images but suffer from flexibility.
Most text-to-image retrieval works focus on the matching of single sentence and single image, which can be classified into global \cite{frome2013devise,kiros2014unifying,wang2016learning,faghri2018vse++,huang2018learning} and dense visual semantic matching models \cite{nam2017dual,karpathy2014deep,karpathy2015deep,lee2018stacked}.
The former employs fixed-dimensional vector as global visual and textual representation. 
Kiros \emph{et al.} \cite{kiros2014unifying} firstly propose to use CNN to encode images and RNN to encode sentences. Faghri \emph{et al.} \cite{faghri2018vse++} propose to mine hard negatives for training.
Sun \emph{et al.} \cite{sun2019videobert} utilize self-supervision from massive instructional videos to learn global visual semantic matching.
Their main limitation is that global vectors are hard to capture fine-grained information.
The dense matching models address such problem via representing image and sentence as a set of fine-grained components.
Nam \emph{et al.} \cite{nam2017dual} sequentially generate image and sentence features, while Karpathy \emph{et al.} \cite{karpathy2015deep} propose to decompose the image as a set of regions and the sentence as a set of words.
The alignment of word and image region not only improves retrieval accuracy, but also makes results more interpretable.
However, few retrieval works have been explored to retrieve image sequences given a story with multiple sentences.
Kim \emph{et al.} \cite{kim2015ranking} deal with long paragraphs but only require short visual summaries corresponding to a common topic.
Ravi \emph{et al.} \cite{ravi2018show} is the closest work to ours to visualize a story with image sequences. 
They propose a coherency model that enhances the single sentence representation with a global hand-crafted coherence vector, and apply global matching to retrieve image for each sentence on visual storytelling dataset VIST \cite{huang2016visual}.
In this work, we not only improve the story-to-image retrieval model via dynamic contextual learning and more interpretable visual semantic dense matching, but also propose an inspire-and-create framework \cite{weston2018retrieve,hashimoto2018retrieve} to improve the flexibility of retrieval-based methods.

\section{Overview of Storyboard Creation}
In this section, we firstly introduce the storyboard creation problem in Section~\ref{sec:problem_intro}, and then describe overall structure of the proposed inspire-and-create framework in Section~\ref{sec:framework}.
Finally, we present our efforts for cinematic image collection in Section~\ref{sec:gm_dataset} which is the foundation to support the inspire-and-create model.

\subsection{Problem Definition}
\label{sec:problem_intro}
Assume $\mathbf{S}=\{S^1, S^2, \cdots, S^N \}$ is a story consisting of $N$ sentences, where each sentence is a word sequence $S^i=\{ w^{i}_{1}, \cdots, w^{i}_{n_i} \}$, the goal of storyboard creation is to generate a sequence of images $\mathbf{I}=\{I^1, I^2, \cdots, I^M\}$ to visualize the story $\mathbf{S}$. The number of images $M$ does not necessarily have to be equal to $N$.

There are two types of training data for the task, called description in isolation (DII) and story in sequence (SIS) respectively.
The DII data only contain pairs of single sentence and single image $(S^i, I^i)$ for training, while the SIS data contain pairs of story and image sequences $(\mathbf{S}, \mathbf{I})$ for training.
However, due to annotation limitations, the $M$ is equal to $N$ in SIS training pairs, which is not always desired in the testing phase.

\subsection{Inspire-and-Create Framework}
\label{sec:framework}
According to the embodied
simulation hypothesis \cite{lakoff2012explaining}, our understanding of language is a simulation based on past experience.
Mimicking such human storyboard creation process, we propose an inspire-and-create framework as presented in Figure~\ref{fig:method_framework}.
It consists of a story-to-image retriever to retrieve existing cinematic images for visual inspiration and a storyboard creator to avoid the inflexibility of using original images via recreating novel storyboard based on the acquired illuminating images.


Specifically, the retriever first selects a sequence of relevant images from existing candidate image set, which are of high-quality and maintain high coverage of details in the story to visualize the story and are employed to inspire the further creator.
Since the candidate images are not specially designed to describe the story, though some regions of images are relevant to the story, there also exist irrelevant regions that should not be presented for interpreting the story.
What is more, the images of high relevancy to the story might not be visually coherent on styles or characters, which can greatly harm human perceptions towards the generated storyboard.
Therefore, the second module, storyboard creator, is proposed to render the retrieved images in order to improve visual-semantic relevancy and visual consistency.

\subsection{Cinematic Image Collection}
\label{sec:gm_dataset}
In order to provide high-quality candidate images for retrieval, we collect a large-scale cinematic image dataset called \textbf{GraphMovie}, which are crawled from a popular movie plot explanatory website\footnote{\url{http://www.graphmovies.com/home/2/}}.

In the website, human annotators sequentially explain events in a movie with a sequence of sentences. Each sentence is aligned with one image extracted from the corresponding period of the event in the movie.
However, the semantic content of the sentence and the image might not be very correlated from visual aspect, because the sentence is an abstract description for the event rather than describing the visual content in the image.
We crawled 8,405 movie explanatory articles from the website which include 3,089,271 aligned sentence and image pairs.
The whole cinematic image set is our retrieval candidate set.
However, it is very inefficient to apply detailed and accurate story-to-image retrieval on the overwhelming size of candidate images.
In order to speed up retrieval, we utilize the aligned sentence to build an initial index for each image.
Given an input sentence query, we first use the whole query or keywords extracted from the query to retrieve top 100 images via the text-text similarity based on this index, which can dramatically reduce the number of candidate images for each sentence.
Then the story-to-image retrieval model is applied on the top 100 images ranked by the text-based retrieval.

\section{Story-to-Image Retriever}
\label{sec:retrieve_model}
There are mainly three challenges to retrieve a sequence of images to visualize a story containing a sequence of sentences.
\textbf{Firstly, sentences in a story are not isolated.} The contextual information from other sentences is meaningful to understand a single sentence. For example, to visualize the following story ``Mom decided to take her daughter to the carnival. They rode a lot of rides. It was a great day!'', contexts from the first sentence are required to understand the pronoun ``they'' in the second sentence as ``mom and daughter''. And overall contexts are beneficial to visualize the third sentence that almost omits visual contents.
\textbf{Secondly, grounding words in image regions is preferred.} The visual grounding not only can improve the retrieval performance via attending to relevant image regions for each word, but also enables future image rendering to erase irrelevant image regions for the story.
\textbf{Finally, the mapping from one sentence to multiple images can be necessary.} Due to constraints of candidate images, sometimes it is hard to employ only one image to represent a detailed sentence, thus retrieving complementary multiple images is necessary. 

In order to address the above challenges, we propose a Contextual-Aware Dense Matching model (CADM) as the story-to-image retriever.
The contextual-aware story encoding is proposed in subsection~\ref{sec:ctx_sent_encoder}  to dynamically employ contexts to understand each word in the story.
In subsection~\ref{sec:dense_match}, we describe the training and inference of dense matching which implicitly learns visual grounding.
Further in subsection~\ref{sec:one_to_many_viz}, we propose a decoding algorithm to retrieve multiple images for one sentence if necessary.

\subsection{Contextual-Aware Story Encoding}
\label{sec:ctx_sent_encoder}

\begin{figure}
	\begin{center}
		\includegraphics[width=1\linewidth]{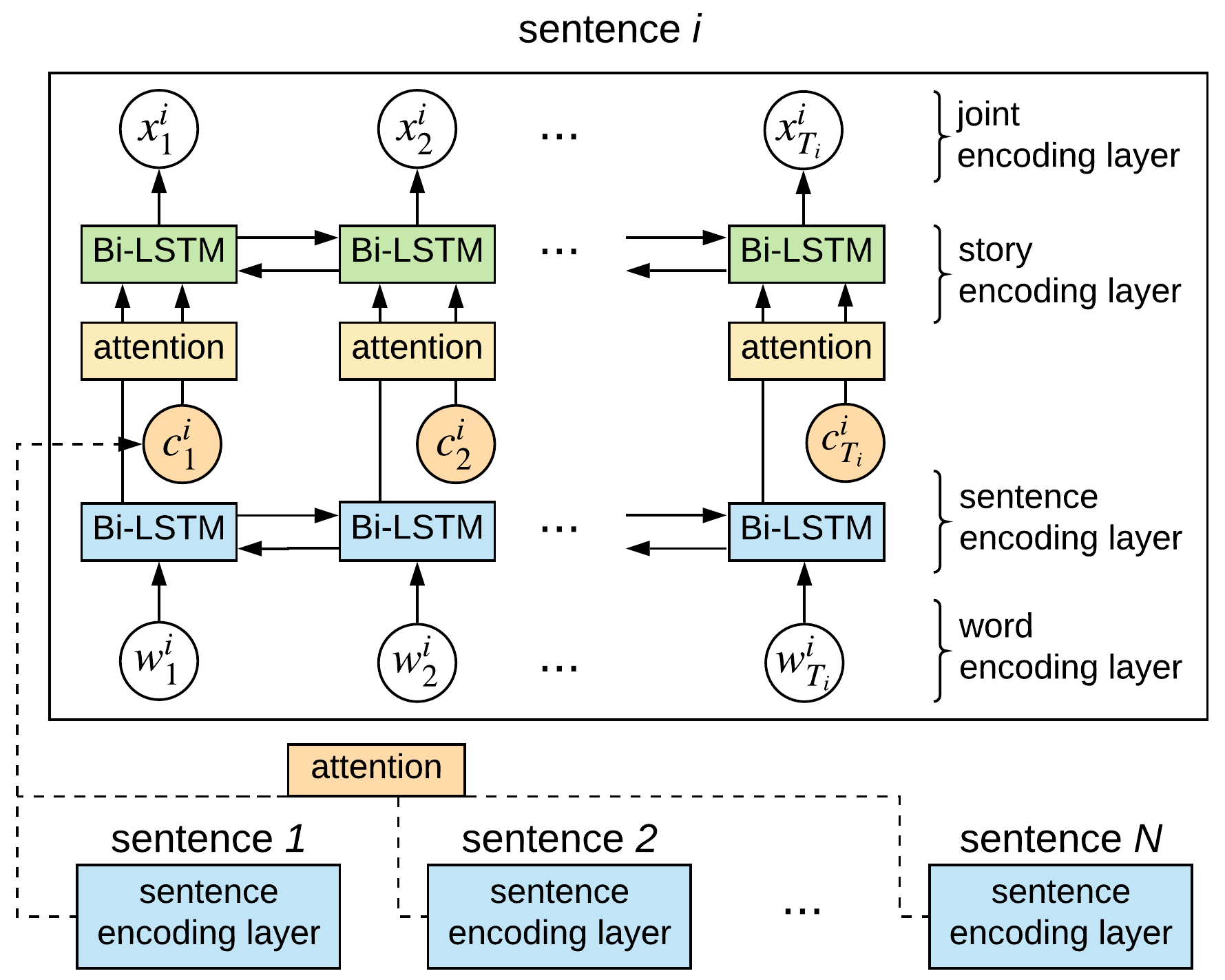}
	\end{center}
	\caption{The contextual-aware story encoding module. Each word is dynamically enhanced with relevant contexts in the story via a hierarchical attention mechanism.}
	\label{fig:ctx_sent_encoder}
\end{figure}

The contextual-aware story encoding dynamically equips each word with necessary contexts within and cross sentences in the story.
As shown in Figure~\ref{fig:ctx_sent_encoder}, it contains four encoding layers and a hierarchical attention mechanism.

The first layer is the word encoding layer, which converts the one-hot word into a distributional vector via word embedding matrix $W_e$.
Then we utilize a bidirectional LSTM \cite{hochreiter1997long} as the second sentence encoding layer to capture contextual information within single sentence $S^i$ for word $w^i_{t}$ as follows:
\begin{equation}
\label{eqn:bilstm}
	\begin{split}
		\overrightarrow{h^i_t} &= \overrightarrow{\mathrm{LSTM}}\ (\overrightarrow{h^i_{t-1}}, W_e w^i_t; \ \overrightarrow{\theta_h}),\\
		\overleftarrow{h^i_t} &= \overleftarrow{\mathrm{LSTM}}\ (\overleftarrow{h^i_{t+1}}, W_e w^i_t; \ \overleftarrow{\theta_h}),\\
		h^i_t &= W_h [\overrightarrow{h^i_t}; \overleftarrow{h^i_t}; W_e w^i_t] + b_h,
	\end{split}
\end{equation}
where $[\cdot]$ is the vector concatenation,  $\overrightarrow{\theta_h}, \overleftarrow{\theta_h}, W_h, b_h$ are parameters to learn, and $h^i_t$ is the word representation equipped with contexts in the sentence for word $w^i_t$.
We can also obtain representation for each single sentence by averaging its constituent word representations:
\begin{equation}
	h^i = \frac{1}{n_i} \sum_{t=1}^{n_i} h^i_t.
\end{equation}

Since the cross-sentence context for each word varies and the contribution of such context for understanding each word is also different, we propose a hierarchical attention mechanism to capture cross-sentence context.
The first attention dynamically selects relevant cross-sentence contexts for each word as follows:
\begin{equation}
	\begin{split}
	    e^i_{t, k} &= v_{a_1} \delta(W_{a_1} [h^i_t; h^k] + b_{a_1}), \\
		a^i_{t, k} &= \mathrm{exp}(e^i_{t, k}) / (\sum_{j=1}^{N} \mathrm{exp}(e^i_{t, j})),\\
		c^i_{t} &= \sum_{k=1}^{N} a^i_{t, k} h^k,
	\end{split}
\end{equation}
where $\delta$ is the nonlinear ReLU function and $v_{a_1}, W_{a_1}, b_{a_1}$ are parameters. 
Given the word representation $h^i_t$ from the second layer and its cross-sentence context $c^i_t$, the second attention adaptively weights the importance of the cross-sentence context for each word:
\begin{equation}
	\begin{split}
		g^i_t &= \sigma(v_{a_2} \delta(W_{a_2} [h^i_t; c^i_t] + b_{a_2})), \\
		\hat{z}^i_t &= g^i_t h^i_t + (1 - g^i_t) c^i_t,
	\end{split}
\end{equation}
where $\sigma$ is the sigmoid function and $v_{a_2}, W_{a_2}, b_{a_2}$ are parameters.
Therefore, $\hat{z}^i_t$ is the word representation equipped with relevant cross-sentence contexts.
To further distribute the updated word representation $\hat{z}^i_t$ within single sentence, we utilize a bidirectional LSTM similar to Eq(\ref{eqn:bilstm}) in the third layer called story encoding, which generates the contextual representation $z^i_t$ for each word.
Finally, in the last layer, we convert $z^i_t$ into the joint visual-semantic embedding space via a linear transformation:
\begin{equation}
	x^i_t = W_x z^i_t + b_x,
\end{equation}
where $W_x, b_x$ are parameters for the linear mapping.
In such way, the $x^i_t$ is encoded with both single sentence and cross sentence context to retrieve images.

\subsection{Dense Visual-Semantic Matching}
\label{sec:dense_match}
In subsection~\ref{sec:ctx_sent_encoder}, we represent sentence $S^i$ as a set of fine-grained word representations $\{x^i_1, \cdots, x^i_{n_i} \}$.
Similarly, we can represent image $I^j$ as a set of fine-grained region representation $\{r^j_1, \cdots, r^j_{m_j} \}$ in the common visual semantic space.
In this work, the image region is detected via a bottom-up attention network \cite{anderson2018bottom} pretrained on the VisualGenome dataset \cite{krishna2017visual}, so that each region represents an object, relation of object or scene.

Based on the dense representations of $S^i$ and $I^j$ and the similarity of each fine-grained cross-modal pair $f(x^i_t, r^j_k)$, we apply dense matching to compute the global sentence-image similarity $F(S^i, I^j)$ as follows:
\begin{equation}
	F(S^i, I^j) = \frac{1}{n_i} \sum_{t=1}^{n_i} \mathrm{max}_{k=1}^{m_j} f(x^i_t, r^j_k),
\end{equation}
where the $f(\cdot)$ is implemented as the cosine similarity in this work.
The dense matching firstly grounds each word with the most similar image region and then averages all word-region similarity over words as the global similarity.

We employ contrastive loss to train the dense matching model, which is:
\begin{equation}
\label{eqn:loss_func}
	L^i = \mathrm{max}(0, \Delta - F(S^i, I^i) + F(S^i, I^j)) + \mathrm{max}(0, \Delta - F(S^i, I^i) + F(S^j, I^i))
\end{equation}
where $(S^i, I^i)$ is matched pair while $(S^i, I^j), (S^j, I^i)$ are mismatched pairs.
The overall loss function is the average of $L^i$ on all pairs in the training dataset.

After training, the dense matching model not only can retrieve relevant images for each sentence, but also can ground each word in the sentence to the most relevant image regions, which provides useful clues for the following rendering.

\begin{algorithm}[tb] 
	\caption{Retrieving a complementary sequence of images to visualize one sentence in a greedy way.} 
	\label{alg:one_to_many_retrieve} 
	\begin{algorithmic}[1] 
		\REQUIRE ~~\\ 
		Sentence $S = \{w_1, \cdots, w_{n}\}$;\\
		Candidate image set $I_c=\{I^1, \cdots, I^{n_c}\}$;

		\ENSURE ~~\\ 
		Selected image sequence $I'_c$;
		
		\STATE Divide $S$ into phrase chunks $\{p_1, \cdots, p_{n_p}\}$ via constituency parsing, where $p_t$ is composed of sequential words; 

		\STATE Computing phrase-image similarity based on the dense matching model  $F(p_t, I^j)=\frac{1}{|p_t|}\sum_{w \in p_t} \max_{r \in I^j} f(w, r)$; 

		\STATE $I'_c = \{\}$ and $I'_r = \{\}$
		\FOR{$t=1$; $t<n_p$; $t++$ }
		\STATE $I^g = \{I^j \mathrm{\ if\ } F(p_t, I^j) \mathrm{\ is\ in\ topK} \}$
		\IF {$I^g \cap I'_r = \emptyset $}
		\STATE $I'_r = I'_r \cup I^g$;
		\STATE $I^t = \mathrm{argmax}_{I^j} F(p_t, I^j)$
		\STATE $I'_c = I'_c \cup \{I^t\}$; 
		\ENDIF
		\ENDFOR
		
		\RETURN $I'_c$; 
	\end{algorithmic}
\end{algorithm}

\subsection{One-to-Many Coverage Enhancement}
\label{sec:one_to_many_viz}

In order to cover as much as details in the story, it is sometimes insufficient to only retrieve one image especially when the sentence is long.
The challenge for such one-to-many retrieval is that we don't have such training data, and whether multiple images are required is dependent on candidate images.
Therefore, we propose a greedy decoding algorithm to automatically retrieve multiple complementary images to enhance the coverage of story contents.

It firstly segments the sentence into multiple phrase chunks via constituency parsing. 
The dense matching model can be used to compute phrase-image similarities for each chunk. 
Then we greedily select top-K images for each chunk because the top-K results are usually similar. 
If the top-K retrieved images for a new chunk haven't been retrieved in previous chunks, it is necessary to visualize the chunk with additional images to cover more details in the sentence.
Otherwise using multiple images can be redundant.
In this greedy decoding way, we can automatically detect the necessity of one-to-many mapping and retrieve complementary images. 
The detailed algorithm is provided in Algorithm~\ref{alg:one_to_many_retrieve}.

\section{Storyboard Creator}
\label{sec:edit_model}

Though retrieved image sequences are cinematic and able to cover most details in the story, they have the following three limitations against high-quality storyboards:
1) there might exist irrelevant objects or scenes in the image that hinders overall perception of visual-semantic relevancy;
2) images are from different sources and differ in styles which greatly influences the visual consistency of the sequence;
and 3) it is hard to maintain characters in the storyboard consistent due to limited candidate images.

In order to alleviate above limitations, we propose the storyboard creator to further refine retrieved images to improve relevancy and consistency.
The creator consists of three modules: 1) automatic relevant region segmentation to erase irrelevant regions in the retrieved image; 2) automatic style unification to improve visual consistency on image styles;  and 3) a semi-manual 3D model substitution to improve visual consistency on characters.

\begin{figure}
	\centering
	\subfigure[Original image.]{ 
		\label{fig:region_seg_origin} 
		\includegraphics[width=0.2\linewidth]{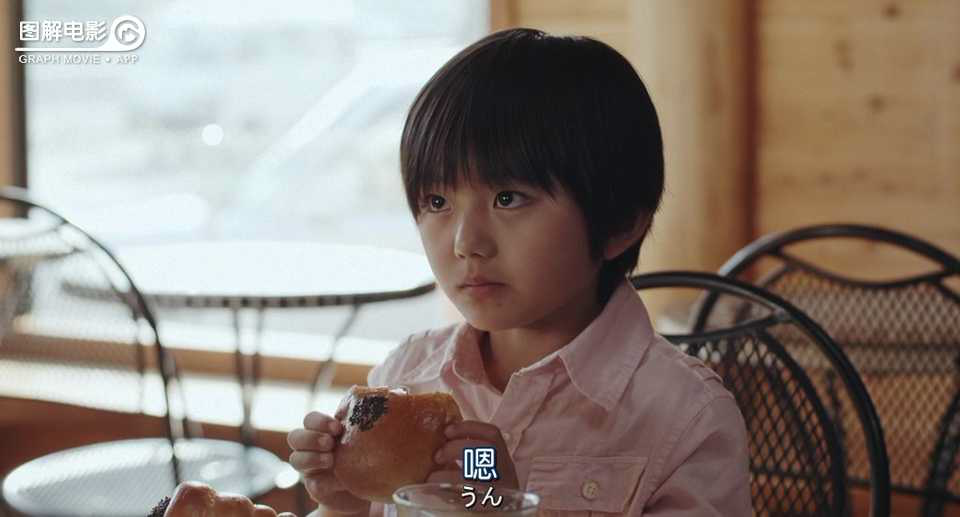}
	} 
	\subfigure[Dense match.]{ 
		\label{fig:region_seg_densematch} 
		\includegraphics[width=0.2\linewidth]{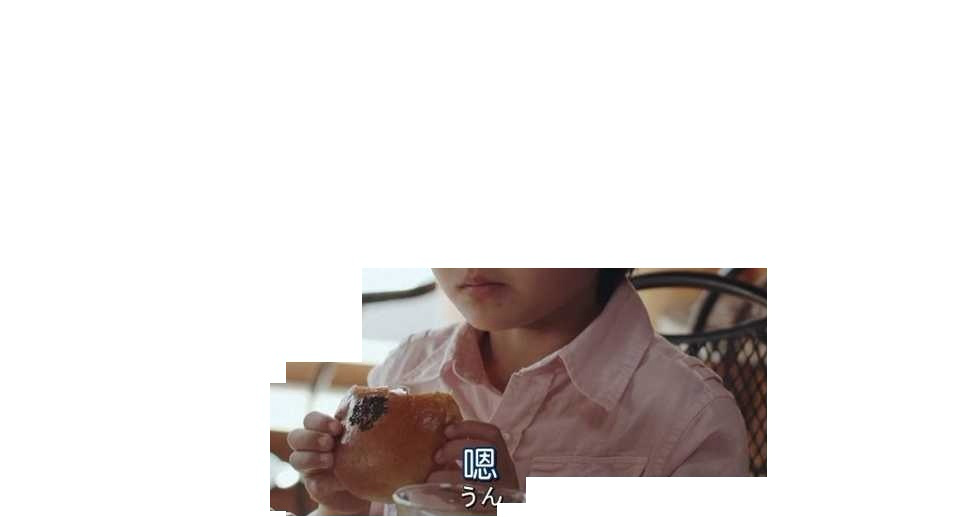}
	} 
	\subfigure[Mask R-CNN.]{ 
		\label{fig:region_seg_maskrcnn} 
		\includegraphics[width=0.2\linewidth]{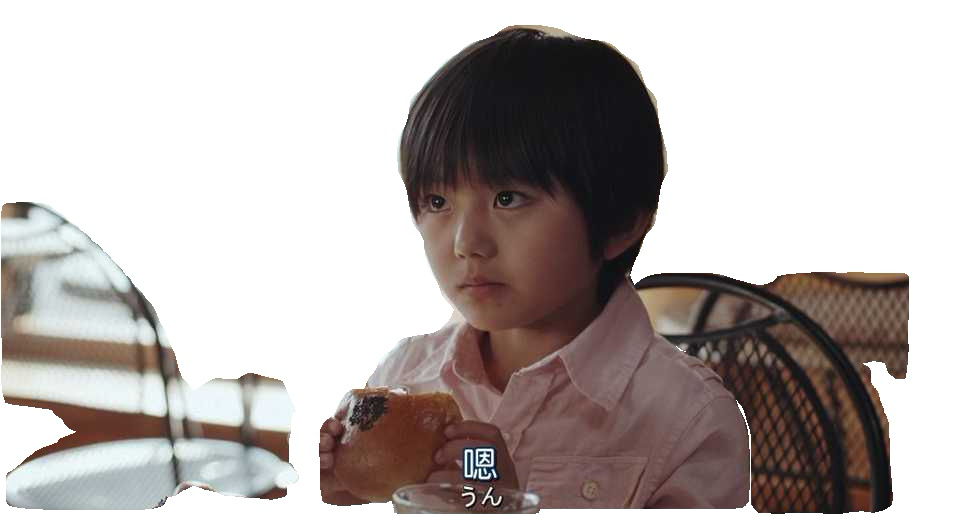}
	} 
	\subfigure[Fusion result.]{ 
		\label{fig:region_seg_fusion} 
		\includegraphics[width=0.2\linewidth]{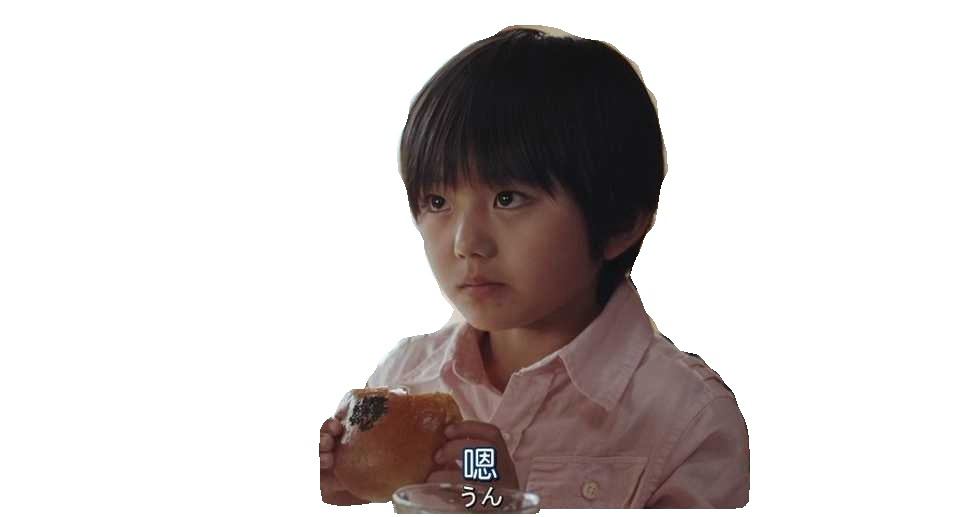}
	} 
	\caption{The dense matching and Mask R-CNN models are complementary for relevant region segmentation.}
	\label{fig:region_segmentation}
\end{figure}

\subsection{Relevant Region Segmentation}
Since the dense visual-semantic matching model grounds each word with a corresponding image region, a naive approach to erase irrelevant regions is to only keep grounded regions.
However, as shown in Figure~\ref{fig:region_seg_densematch}, although grounded regions are correct, they might not precisely cover the whole object because the bottom-up attention \cite{anderson2018bottom} is not especially designed to achieve high segmentation quality.
The current Mask R-CNN model \cite{he2017mask} is able to obtain better object segmentation results. However it cannot distinguish the relevancy of objects and the story in Figure~\ref{fig:region_seg_maskrcnn}, and it also cannot detect scenes.
Since these two methods are complementary to each other, we propose a heuristic algorithm to fuse the two approaches to segment relevant regions precisely. 

For each grounded region by the dense matching model, we align it with a object segmentation mask from the Mask R-CNN model.
If the overlap between the grounded region and the aligned mask is bellow certain threshold, the grounded region is likely to be relevant scenes.
In such case, we keep the grounded region.
Otherwise the grounded region belongs to an object and we utilize the precise object boundary mask from Mask R-CNN to erase irrelevant backgrounds and complete relevant parts.
As shown in Figure~\ref{fig:region_seg_fusion}, the proposed fusion method improves the separate processing model and overall image relevancy.

\subsection{Style Unification}
We explore style transfer techniques to unify the styles of retrieved image sequences.
Specifically, we convert the real-world images into cartoon style images.
On one hand, the cartoon style images maintain the original structures, textures and basic colors, which ensures the advantage of being cinematic and relevant.
On the other hand, the cartoon style is more simplified and abstract than the realistic style, so that such cartoon sketch can improve visual consistency of the storyboard.
In this work, we utilize a pretrained CartoonGAN \cite{chen2018cartoongan} for the cartoon style transfer.

\subsection{3D Character Substitution}
It is very challenging to maintain visual consistency on characters of the automatically created storyboard.
Therefore, we propose a semi-manual way to address this problem, which involves manual assistance to improve the character coherency.

Since above created storyboard has presented good structure and organization of character and scenes, we refer to manual effort to mimic the visualization of automatic created storyboard by changing the scene and characters into some predefined templates.
The templates are from a 3D software Autodesk Maya 
, which can be easily dragged and organized in the Maya software to construct a 3D picture.
The human storyboard artist is asked to select proper templates to replace the original ones in the retrieved image.
The character is placed according to its action and location in the retrieved image, which can greatly reduce the costly human design. 
After the placement of all characters and scenes, we can select a camera view to render the 3D model to 2D image.
Although currently this step still requires some human efforts, we will make it more automatic in our future work.

\section{Experiment}

\subsection{Training Datasets}

\noindent
\textbf{MSCOCO}:
The MSCOCO \cite{lin2014microsoft} dataset belongs to the DII type of training data.
It consists of 123,287 images and each image is annotated with 5 independent single sentences. 
Since the MSCOCO cannot be used to evaluate story visualization performance, we utilize the whole dataset for training.

\noindent
\textbf{VIST}: The VIST dataset is the only currently available SIS type of dataset.
Following the standard dataset split \cite{huang2016visual}, we utilize 40,071 stories for training, 4,988 stories for validation and 5,055 stories for testing.
Each story contains 5 sentences as well as the corresponding ground-truth images.

\subsection{Testing Stories in the Wild}
To evaluate the storyboard creation performance for stories in the wild, we collect 200 stories from three different sources including Chinese idioms, movie scripts and sentences in the GraphMovie plot explanations, which contain 579 sentences in total.
For each sentence, we manually annotate its relevancy with the top 100 candidate images from the text-based retrieval as explained in subsection~\ref{sec:gm_dataset}.
An image is relevant to a sentence if it can visualize some parts of the sentence.
We find that only 62.9\% (364 in 579) sentences contain relevant images in the top 100 candidates by text retrieval, which demonstrates the difficulty of the task.
There are 10.9\% relevant images on average for those 62.9\% sentences.
We have released the testing dataset\footnote{ \url{https://github.com/cshizhe/neuralstoryboard}}.

\subsection{Experimental Setups}
We combine the MSCOCO dataset and VIST training set for training, and evaluate the story-to-image retrieval model on VIST testing set and our GraphMovie testing set respectively.
Since the GraphMovie dataset is in Chinese, we translate the training set from English to Chinese via an automatic machine translation API \footnote{\url{https://www.bing.com/translator}}.
In order to cover out-of-vocabulary words besides the training set, we fix the word embeddings with pretrained word vectors in the retrieval model, with Glove embedding \cite{pennington2014glove} for English and fastText embedding \cite{grave2018learning} for Chinese.
The dimensionality of joint embedding space is set as 1024.
Hard negative mining is applied to select negative training pairs within a mini-batch \cite{faghri2018vse++} in Eq~(\ref{eqn:loss_func}) for MSCOCO dataset.
However, since the sentences in the VIST stories are more abstract than caption descriptions in MSCOCO, one story sentence can be suitable for different images which makes the selected ``hard negatives'' noisy.
Therefore, we average over all negatives for the VIST dataset to alleviate noisy negatives.
The model is trained by Adam algorithm with learning rate of 0.0001 and batch size of 256.
We train the model for 100 epochs and select the best one according to the performance on VIST validation set.

To make fair comparison with the previous work \cite{ravi2018show}, we utilize the Recall@K (R@K) as our evaluation metric on VIST dataset, which measures the percentage of sentences whose ground-truth images are in the top-K of retrieved images.
We evaluate under $K=10, 50$ and $100$ as in \cite{ravi2018show}.
For the GraphMovie testing dataset, since the number of candidate images for each sentence is less than 100, we evaluate R@K with K=1, 5, and 10.
We also evaluate on common retrieval metrics including median rank (MedR), mean rank (MeanR) and mean average precision (MAP).

\begin{table}[t]
	\centering
	\caption{Story-to-image retrieval performance on the VIST testing set. All scores are reported as percentage (\%).}
	\label{tab:vist_results}
	\begin{tabular}{c|ccc} \toprule
		& R@10 & R@50 & R@100 \\ \midrule
		CNSI \cite{ravi2018show} & 0 & 1.5 & 4.5 \\
		No Context \cite{karpathy2015deep} & 11.24 & 28.38 & 39.15 \\ \midrule
		CADM w/o attn & 12.98 & 32.84 & 44.47 \\
		\textbf{CADM} & \textbf{13.65} & \textbf{33.91} & \textbf{45.53} \\ \bottomrule
	\end{tabular}
\end{table}

\subsection{Quantitative Results}
We first evaluate the story-to-image retrieval performance on the in-domain dataset VIST.
We compare our CADM model with two state-of-the-art baselines and one variant:
\begin{itemize}
	\item CNSI \cite{ravi2018show}: global visual semantic matching model which utilizes hand-crafted coherence feature as encoder.
	\item No Context \cite{karpathy2015deep}: the state-of-the-art dense visual semantic matching model for text-to-image retrieval.
	\item CADM w/o attn: variant of CADM model, which does not use attention to dynamically compute context but averages representations of other sentences as fixed context.
\end{itemize}
Table~\ref{tab:vist_results} presents the story-to-image retrieval performance of the four models on VIST testing set.
The ``No Context'' model has achieved significant improvements over the previous CNSI \cite{ravi2018show} method, which is mainly contributed to the dense visual semantic matching with bottom-up region features instead of global matching.
The CADM model without attention can boost the performance of ``No Context'' model with fixed context, which demonstrates the importance of contextual information for the story-to-image retrieval.
Our proposed CADM model further achieves the best retrieval performance because it can dynamically attend to relevant story context and ignore noises from context.

\begin{table}[t]
	\centering
	\caption{Story-to-image retrieval performance on the GraphMovie testing set. All scores are reported as percentage (\%).}
	\label{tab:gm_results}
	\begin{tabular}{c|cccccc} \toprule
		& R@1 & R@5 & R@10 & MedR & MeanR & MAP \\ \midrule
		Text & 26.37 & 49.73 & 62.64 & 5.0 & 13.6 & 24.6 \\ \midrule
		No Context & 25.00 & 53.30 & 65.66 & 4.0 & 11.1 & 26.4 \\
	    CADM & 26.65 & 54.95 & 67.03 & 3.0 & 10.4 & 27.6  \\ \midrule
		\multicolumn{1}{c|}{CADM+Text} & \textbf{32.97} & \textbf{61.26} & \textbf{74.73} & \textbf{2.0} & \textbf{7.9} & \textbf{31.8} \\ \bottomrule
	\end{tabular}
\end{table}

Then we explore the generalization of our retriever for out-of-domain stories in the constructed GraphMovie testing set.
We compare the CADM model with the text retrieval based on paired sentence annotation on GraphMovie testing set and the state-of-the-art ``No Context'' model.
As shown in Table~\ref{tab:gm_results}, the purely visual-based retrieval models (No Context and CADM) improve the text retrieval performance since the annotated texts are noisy to describe the image content.
Among visual-based retrieval models, the proposed CADM model also outperforms ``No Context'' model for out-of-domain stories, which further demonstrates the effectiveness of the proposed model even in more difficult scenarios.
It achieves median rank of 3, which indicates there exist one relevant image in the top 3 retrieved ones for most story sentences.
Our CADM model is also complementary with the text retrieval as seen in the last row of Table~\ref{tab:gm_results}.
When we fuse the visual retrieval scores and text retrieval scores with 0.9 and 0.1 weights respectively, we achieve the best performance on the GraphMovie testing set.

\begin{table}
	\centering
	\caption{Story-to-image retrieval performance on subset of GraphMovie testing set that only includes Chinese idioms or movie scripts. All scores are reported as percentage (\%).}
	\label{tab:gm_outdomain_results}
	\begin{tabular}{c|cccccc} \toprule
		& R@1 & R@5 & R@10 & MedR & MeanR & MAP \\ \midrule
		Text & 14.01 & 36.71 & 51.21 & 9.0 & 18.2 & 17.4 \\ \midrule
		No Context & 21.26 & 48.31 & 61.35 & 5.0 & 12.5 & 22.9 \\ 
		CADM & 22.22 & 51.21 & 65.22 & 4.0  & 10.5 & 25.5 \\ \midrule
		\multicolumn{1}{c|}{CADM+Text} & \textbf{25.60} & \textbf{51.21} & \textbf{67.15} & \textbf{4.0} & \textbf{9.9} & \textbf{26.5} \\ \bottomrule
	\end{tabular}
\end{table}

Since the GraphMovie testing set contains sentences from text retrieval indexes, it can exaggerate the contributions of text retrieval.
Therefore, in Table~\ref{tab:gm_outdomain_results} we remove this type of testing stories for evaluation, so that the testing stories only include Chinese idioms or movie scripts that are not overlapped with text indexes.
We can see that the text retrieval performance significantly decreases compared with Table~\ref{tab:gm_results}.
Nevertheless, our visual retrieval performance are almost comparable across different story types, which indicates that the proposed visual-based story-to-image retriever can be generalized to different types of stories.

\begin{table}
\centering
\caption{Human evaluation performance on 30 selected Chinese idioms. All scores are reported as percentage (\%).}
\label{tab:human_evaluation}
\begin{tabular}{c|ccc} \toprule
 & good & so-so & bad \\ \midrule
No Context & 38.4 & 36.0 & 25.6 \\
CADM & \textbf{52.2} & 25.4 & 22.4 \\ \bottomrule
\end{tabular}
\end{table}

\begin{figure*}
	\begin{center}
		\includegraphics[width=1\linewidth]{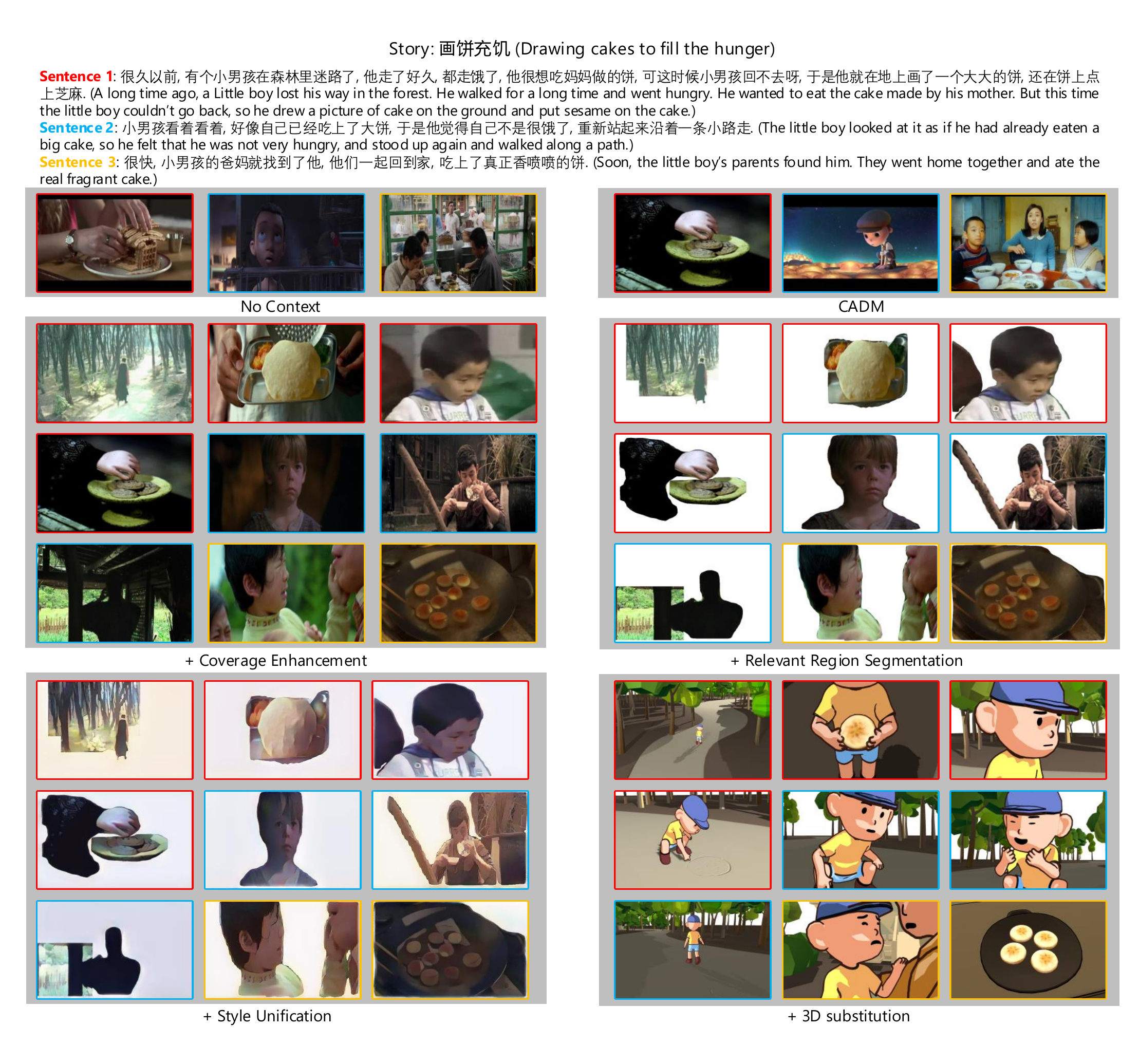}
	\end{center}
	\vspace{-6mm}
	\caption{Example of one story visualization of different retrieval models and rendering steps for an out-of-domain story. Images of different colors are corresponding to sentences with the same color in the story. Best viewed in colors.}
	\label{fig:visualization}
\vspace{-2mm}
\end{figure*}

\subsection{Qualitative Results}
Due to the subjectivity of the storyboard creation task, we further conduct human evaluation on the created storyboard besides the quantitative performance.
We randomly select 30 Chinese idioms including 157 sentences for the human evaluation.
The ``No Context'' retrieval model and the proposed CADM model are utilized to select top 3 images for each story respectively.
We ask human annotators to classify the best of the top 3 retrieved images for each sentence into 3 level based on the visualization quality: good, so-so and bad.
Each sentence is annotated by three annotators and the majority vote is used as the final annotation.
Table~\ref{tab:human_evaluation} presents the human evaluation results of the two models.
The CADM achieves significantly better human evaluation than the baseline model.
More than 77.6\% of sentences are visualized reasonably according to the user study (good and so-so).

In Figure~\ref{fig:visualization}, we provide the visualization of different retrievers and rendering processes of the proposed inspire-and-create framework for a Chinese idiom ``Drawing cakes to fill the hunger''.
The contextual-aware CADM retriever selects more accurate and consistent images than the baseline especially for the third sentence in this case.
The proposed greedy decoding algorithm further improves the coverage of long sentences via automatically retrieving multiple complementary images from candidates.
The relevant region segmentation rendering step erases irrelevant backgrounds or objects in the retrieved image, though there exist minor mistakes in the automatic results.
And through style unification we can obtain more visually consistent storyboard.
The last row is the manually assisted 3D model substitution rendering step, which mainly borrows the composition of the automatic created storyboard but replaces main characters and scenes to templates.
The final human assisted version is nearly comparable to professional storyboard.
We will make this step more automatic in our future work.
In general, we managed to integrate the whole system into a commercial product that tells visual stories to children.

\section{Conclusion}
In this work, we focus on a new multimedia task of storyboard creation, which aims to generate a sequence of images to illustrate a story containing multiple sentences.
We tackle the problem with a novel inspire-and-create framework, which includes a story-to-image retriever to select relevant cinematic images for vision inspiration and a creator to further refine images and improve the relevancy and visual consistency.
For the retriever, we propose a contextual-aware dense matching model (CADM), which dynamically employs contextual information in the story with hierarchical attentions and applies dense visual-semantic matching to accurately retrieve and ground images.
For the creator, we propose two fully automatic rendering steps for relevant region segmentation and style unification and one semi-manual steps to substitute coherent characters.
Extensive experimental results on in-domain and out-of-domain visual story datasets demonstrate the effectiveness of the proposed inspire-and-create model.
We achieve better quantitative performance in both objective and subjective evaluation than the state-of-the-art baselines for storyboard creation, and the qualitative visualization further verifies that our approach is able to create high-quality storyboards even for stories in the wild.
In the future, we plan to recognize the picturing angles of images and learn the way of placing them in movies and guide the neural storyboard artist to create more professional and intelligent storyboards.

\begin{acks}
The authors would like to thank Qingcai Cui for cinematic image collection, Yahui Chen and Huayong Zhang for their efforts in 3D character substitution.
This work was supported by National Natural Science Foundation of China (No. 61772535), Beijing Natural Science Foundation (No. 4192028), and National Key Research and Development Plan (No. 2016YFB1001202).
\end{acks}

\bibliographystyle{unsrt}
\balance
\bibliography{reference}

\end{document}